\documentclass{article}
\usepackage{spconf,amsmath,graphicx}

\usepackage{enumitem}
\setlist{nosep, leftmargin=14pt}
\usepackage[colorlinks=true, linkcolor=black, urlcolor=black, citecolor=black]{hyperref}
\usepackage[capitalise,nameinlink,noabbrev]{cleveref}

\usepackage{mwe} 

\title{MIRAGE: Multimodal Identification and Recognition of Annotations in Indian
General Prescriptions}
\name{Tavish Mankash$^{1}$ \quad V. S. Chaithanya Kota$^{2}$ \quad Anish De$^{3}$ \quad Praveen Prakash$^{4}$ \quad Kshitij Jadhav$^{5}$}

\address{
   $^{1}$ \href{mailto:tavish.mankash@gmail.com}{tavish.mankash@gmail.com} 
   $^{2}$ \href{mailto:kotachaithanya1@gmail.com}{kotachaithanya1@gmail.com} \\
   $^{3}$ \href{mailto:anishde2007@gmail.com}{anishde2007@gmail.com} 
   $^{4}$ \href{mailto:praveen@mtatva.com}{praveen@mtatva.com} 
   $^{5}$ \href{mailto:kshitij.jadhav@iitb.ac.in}{kshitij.jadhav@iitb.ac.in}
}

\begin{document}

\maketitle

\begin{abstract}
  Hospitals in India still rely on handwritten medical records despite the availability of Electronic Medical Records (EMR), complicating statistical analysis and record retrieval. Handwritten records pose a unique challenge, requiring specialized data for training models to recognize medications and their recommendation patterns. While traditional handwriting recognition approaches employ 2-D LSTMs, recent studies have explored using Multimodal Large Language Models (MLLMs) for OCR tasks. Building on this approach, we focus on extracting medication names and dosages from simulated medical records. Our methodology \textbf{MIRAGE} (\textbf{M}ultimodal \textbf{I}dentification and \textbf{R}ecognition of \textbf{A}nnotations in Indian \textbf{GE}neral Prescriptions) involves fine-tuning the \textit{QWEN VL}, \textit{LLaVA 1.6} and \textit{Idefics2} models on \textit{743,118 high-resolution simulated medical record images - fully annotated} from \textit{1,133 doctors} across India. Our approach achieves \textit{82\%} accuracy in extracting medication names and dosages. 

\end{abstract}

\section{Introduction}
\label{sec:introduction}

Handwritten prescriptions dominate medical records in India, despite known high error rates. Once written, they are often impossible for untrained individuals to decipher without the help of a pharmacist. A study conducted in South Africa concluded that pharmacists read medical records with a median correct percentage of 75\% indicating the increased risk of wrongful medication \cite{brits2017illegible}.

\begin{table}[hb]
\centering

\begin{tabular}{|l|c|c|c|}
\hline
\textbf{Model} & \textbf{Precision (\%)} & \textbf{Recall (\%)} & \textbf{F1 (\%)} \\
\hline
LLaVA 1.6          & 2.00                   & 2.00                 & 2.00                   \\
Gemini 1.5 Pro  & 5.65                   & 5.46                 & 5.53                   \\
GPT 4o          & 7.57                   & 7.57                 & 7.57                   \\
\hline
\end{tabular}
\caption{Scores of various LLMs on our dataset without fine-tuning.}
\label{tab:top_base}
\end{table}

Addressing the challenge of accurately reading handwritten prescriptions is complex and cannot be solved effectively using unspecialized models (see \Cref{tab:top_base}). MLLMs have recently demonstrated state-of-the-art performance in Optical Character Recognition (OCR) \cite{kim2022ocr}. Building on this success, we applied these models to the challenging task of Hand Writing Recognition (HWR). Our results are promising and in real world scenarios surpass existing computer vision approaches.

\section{Literature Review}
\label{sec:format}

\subsection{HWR with LLMs}

Fadeeva \etal used MLLMs for online handwriting recognition (not paper based), achieving state-of-the-art accuracy by representing handwriting data with various methods, including color coding for step size \& duration \cite{fadeeva2024representing}. Additionally, another study that investigated the performance of MLLMs in reading tasks notes HWR and semantic reliance as two primary weaknesses \cite{liu2023hidden}. 

\subsection{Reading Handwritten Prescriptions with AI}

\begin{figure}[h!]
    \centering
    \includegraphics[width=\columnwidth]{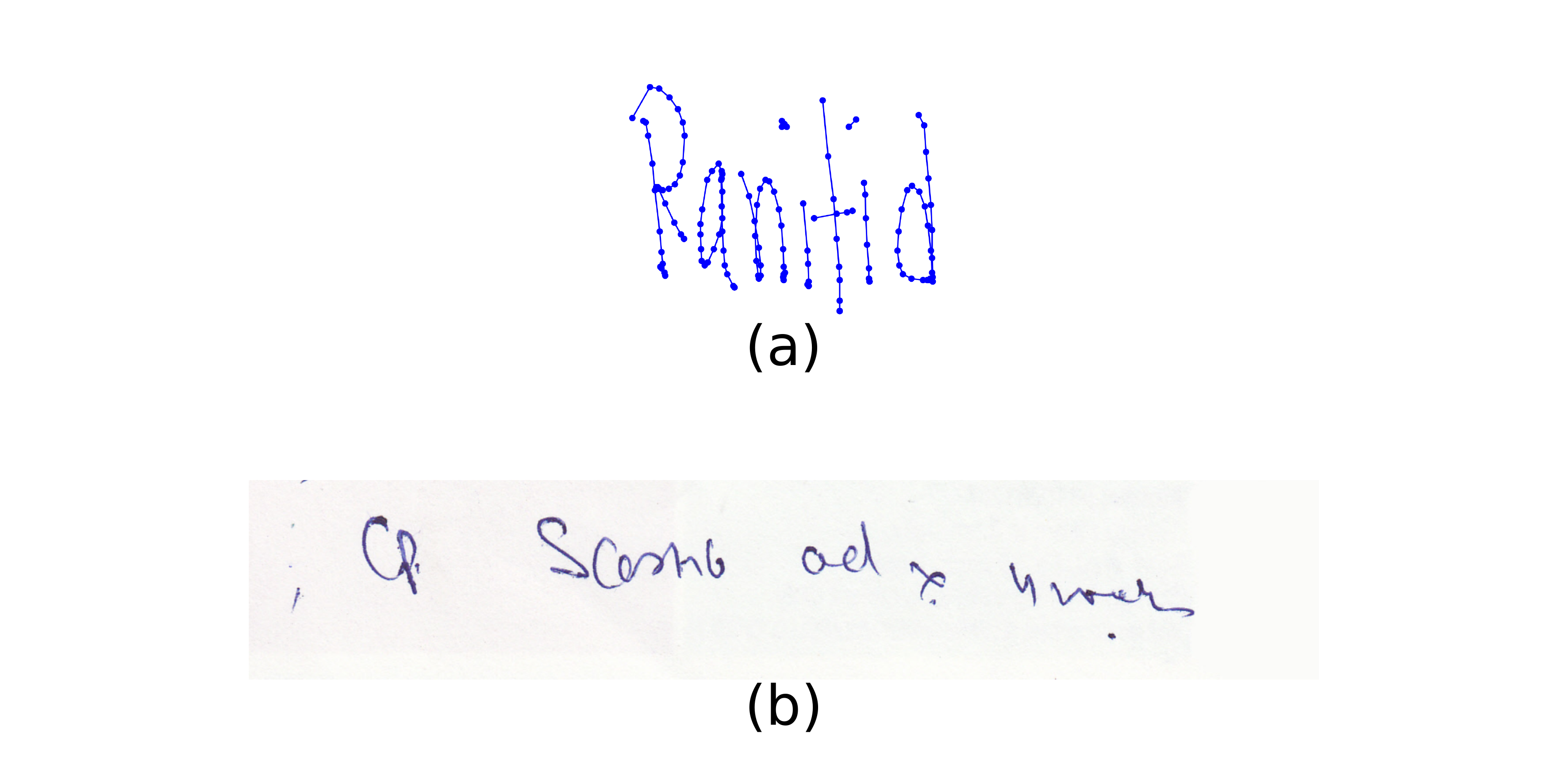}
    \caption{Difference between digital and paper-written handwriting. (a) From \cite{tabassum2021recognition} by Tabassum \etal, reprinted with permission © 2021 IEEE (b) Typical prescribed medication from our dataset. }
    \label{fig:clean-vs-genuine}
\end{figure}

Reported accuracies for handwritten prescription recognition remain impractical for real-world use. One study achieved an accuracy of 64-70\% with a model first trained on the IAM dataset, while another reached 74.13\% using a lexicon-driven approach, which was also limited by its small 520 word lexicon \cite{9357136, chumuang2018handwritten}. Another work performing online recognition achieved 68.78\% accuracy averaged across all users \cite{rani2022recognition}. However, this study had only 9 users from whom data was collected with a total of 3000 medicine samples. While Tabassum \etal report 89\% accuracy for the simpler online handwritten medical word recognition using BLSTMs and SRP augmentation, their system suffers from the following limitations: a small, unrealistic dataset comprising of 17,431 handwritten samples of 480 medical words from 39 doctors; samples were collected on a tablet resulting in artificially clean handwriting—see \cref{fig:clean-vs-genuine}; impractical for developing countries due to its stylus requirement \cite{tabassum2021recognition}.

\begin{figure}[ht!]

\begin{minipage}[b]{1.0\linewidth}
  \centering
  \centerline{\includegraphics[width=8.5cm]{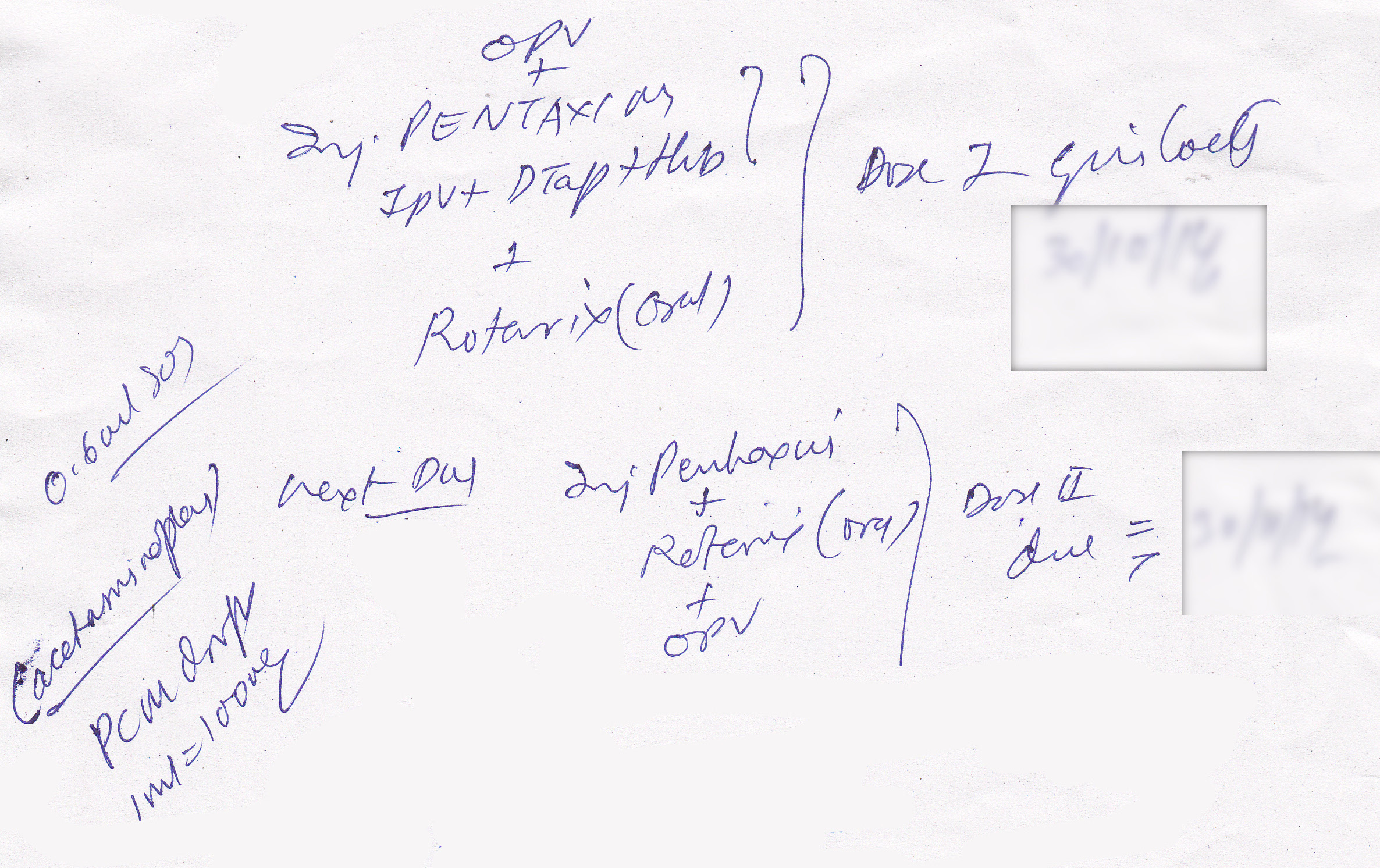}}
  \centerline{(a) Sections, non-text elements and diagonal writing}\medskip
\end{minipage}
\begin{minipage}[b]{.48\linewidth}
  \centering
  \centerline{\includegraphics[width=4.0cm]{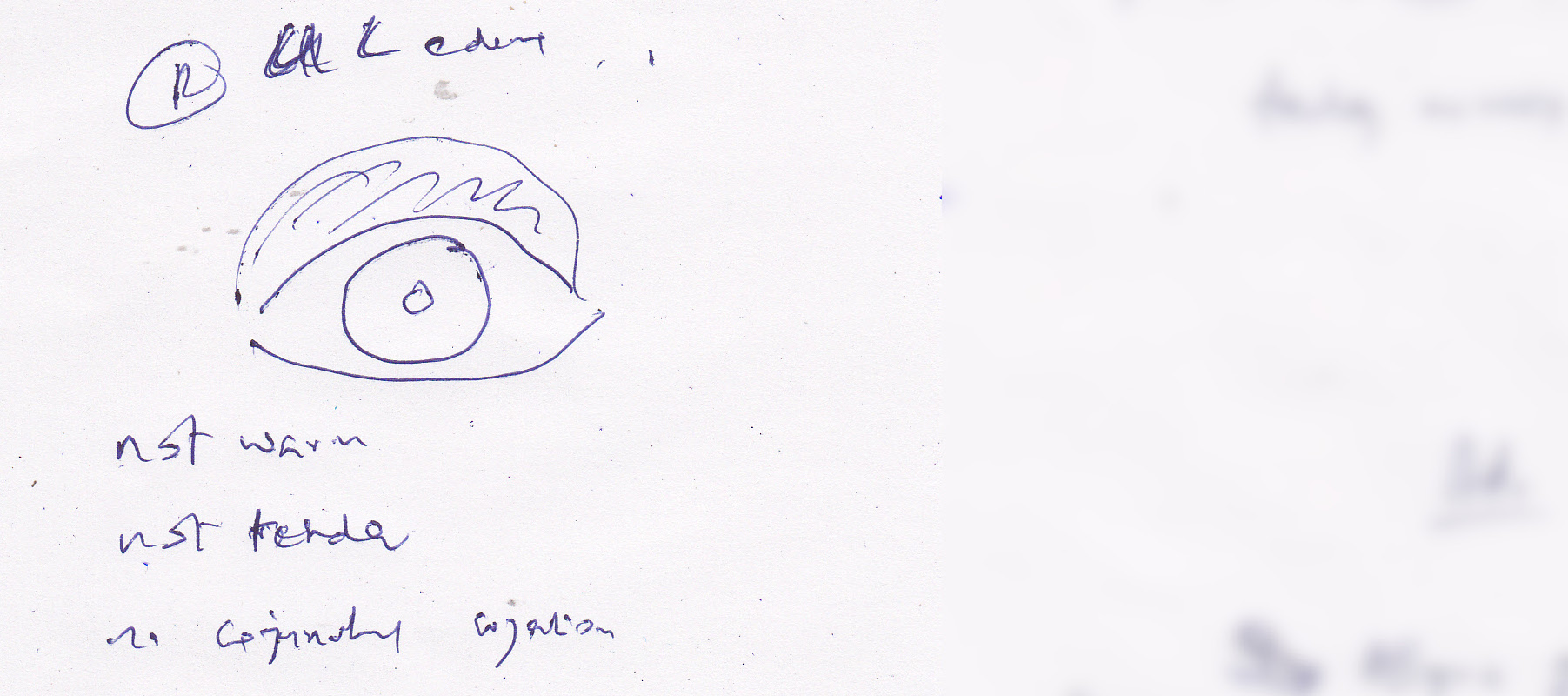}}
  \centerline{(b) Eye diagram}\medskip
\end{minipage}
\hfill
\begin{minipage}[b]{0.48\linewidth}
  \centering
  \centerline{\includegraphics[width=4.0cm]{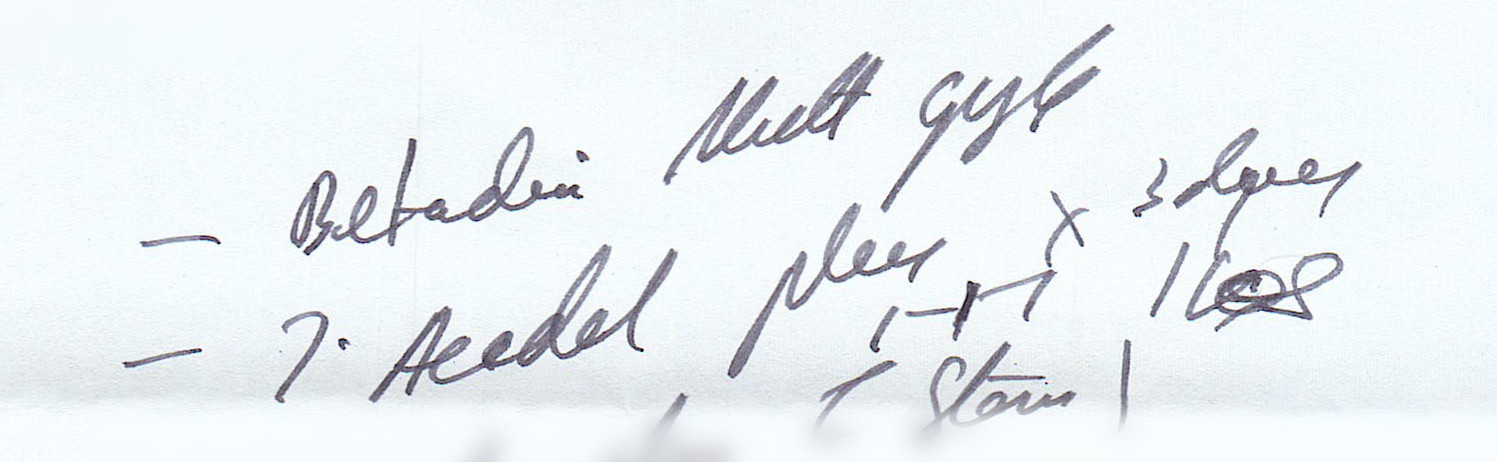}}
  \centerline{(c) Unclear strokes }\medskip
\end{minipage}
\caption{(a) Tilted text complicates word isolation. Chronological order would be incomprehensible without understanding sectioning}
\label{fig:combined}
\end{figure}

Current handwritten prescription recognition models are limited by: inadequate datasets—trained primarily on small, unrepresentative prescription datasets, these models struggle to capture the real-world variability of handwriting and lack the diversity needed to detect rarer medications; insufficient training on medical abbreviations \cite{9357136, chumuang2018handwritten, rani2022recognition}—this contributes to 60\% of medication name errors by experts\cite{brits2017illegible}; insufficient attention to crucial non-textual and dosage information, as in \cref{fig:combined} (a) \& (b) \cite{9357136, chumuang2018handwritten, rani2022recognition, tabassum2021recognition}; and a word-level segmentation approach also suboptimally handles complex prescriptions, as illustrated in \cref{fig:combined} (a) \cite{9357136, chumuang2018handwritten, rani2022recognition, tabassum2021recognition}. Our work naturally handles non-textual, dosage and abbreviations information as we feed the entire medical record to the LLM (thus, we don't do word-level splitting).

\section{Methodology}
\label{sec:methodology}

\begin{figure}[ht]
    \begin{minipage}{0.48\linewidth}
        \centering
        \includegraphics[width=\linewidth]{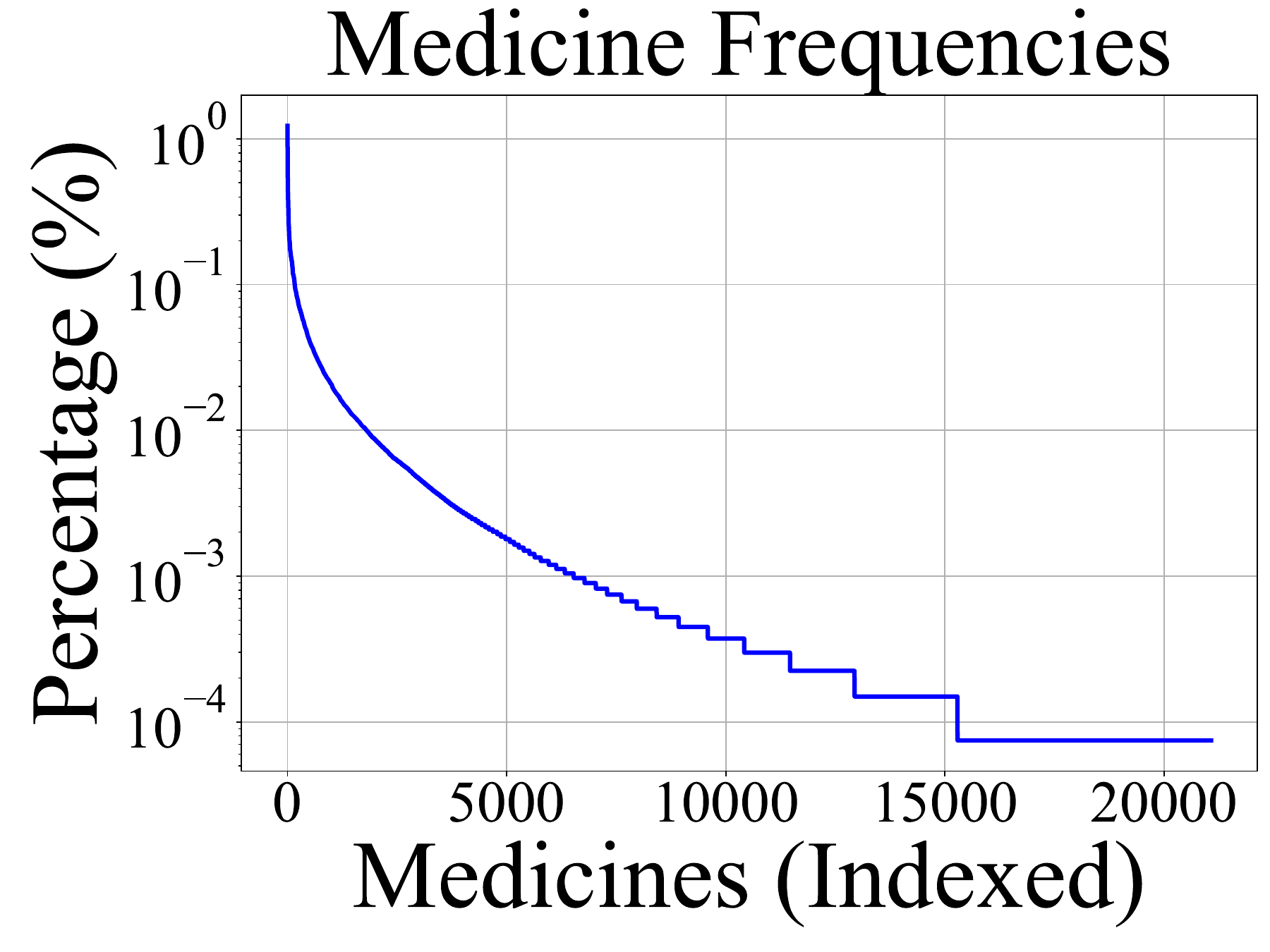}
        \caption{Medicine name distribution, with the most frequent on the left. Percentage frequency is in log scale.}
        \label{fig:med-freq}
    \end{minipage}
    \hfill
    \begin{minipage}{0.48\linewidth}
        \centering
        \includegraphics[width=\linewidth]{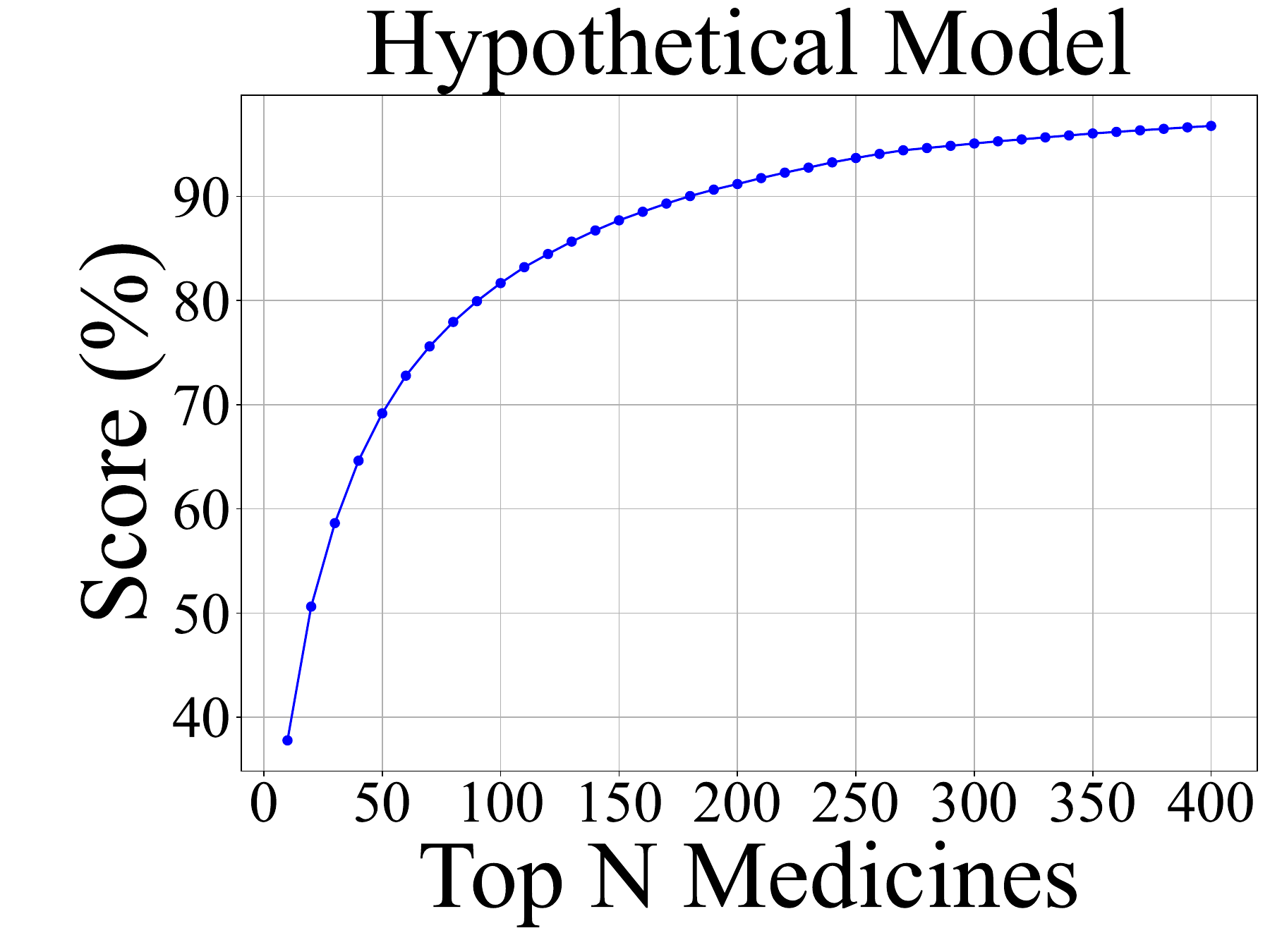}
        \caption{Model precision when identifying the top 'N' most frequently prescribed medications per doctor.}
        \label{fig:top_n}
    \end{minipage}
\end{figure}

\textit{\textbf{Dataset:}}
The novelty of our work lies in the application of MLLMs and the utilization of a novel simulated dataset of 743,118 handwritten medical records, realistically mimicking patient type frequencies, created by 1,133 doctors across 52 specialties with the 6 most frequent shown in \cref{tab:specialties}. After filtering for spam, 513,407 records were used for training and 15,000 for testing. This dataset includes 1,386,015 individual medicines prescribed from a pool of 21,075 unique medicine brand names. Rare prescriptions are notably common (see \cref{fig:med-freq}). In a simulation we conducted, a hypothesized model identifies the top 'N' most frequently prescribed medications by each doctor, aiming to enhance recognition precision by prioritizing common medicines and ignoring those prescribed less frequently, illustrating the increasing importance of detecting these rarer medications for achieving higher accuracies (see \cref{fig:top_n}). A subset of 100 prescriptions has been made publicly available \cite{100dataset}.

\begin{table}[ht]
    \centering
    \begin{tabular}{|l|c|}
        \hline
        \textbf{Specialty} & \textbf{Number of Prescriptions} \\
        \hline
        Physician & 79,676 \\
        Pediatrician & 68,420 \\
        Neurologist & 49,573 \\
        Gynecologist & 48,388 \\
        Not Mentioned & 43,633 \\
        Cardiologist & 37,385 \\
        \hline
    \end{tabular}
    \caption{Frequency distribution of medical records across various medical specialties in the dataset.}
    \label{tab:specialties}
\end{table}

\textit{\textbf{Extracting all information:}} We initially aimed to extract all information (age, gender, vitals, medication dosages and frequencies, diagnostics, and diagnoses) from the simulated medical records. To achieve this, we fine-tuned \textit{QWEN VL} and LLaVA \cite{bai2023qwenvlversatilevisionlanguagemodel, liu2024llavanext}. We also evaluated the impact of dataset size and character spacing (e.g., M O N T A I R  F X) within the targets to reduce semantic reliance, as suggested in \cite{fadeeva2024representing}. The training process took 9 days, utilizing 7 \textit{A6000} GPUs.

\begin{figure}
    \centering
    \includegraphics[width=1\linewidth]{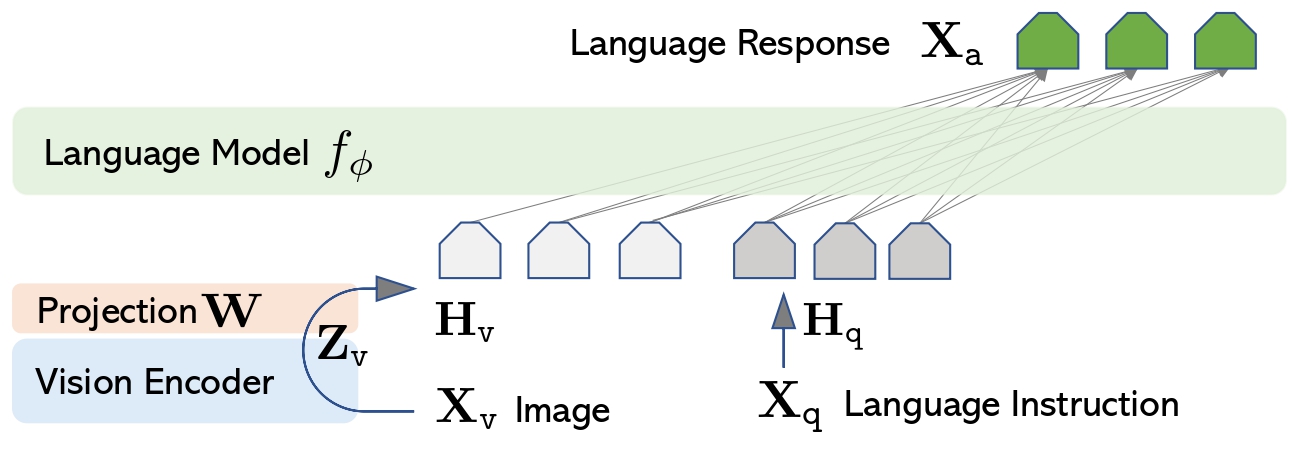}
    \caption{LLaVA architecture. Source: Liu \etal \cite{liu2023llava}}
    \label{fig:liu-llava-arch}
\end{figure}

\textit{\textbf{Medicine names and dosages with LLaVA:}} Owing to low performance while extracting all details from these simulated medical records, we settled on the less ambitious goal of only extracting medicine names. We trained LLaVA 1.6 with the Mistral 7B LLM and the CLIP vision encoder, connected via a trainable projector as in \cref{fig:liu-llava-arch} \cite{liu2024llavanext, radford2021learningtransferablevisualmodels}. We employed LoRA for fine-tuning with a rank of 128 and an alpha of 256 to allow extensive fine-tuning. The initial learning rate was set at 2e-5, with a warm-up ratio of 0.03. We also used DeepSpeed’s ZeRO Stage 3 to train on 7 A6000s for 3.5 days. In our experiments, the learning rate declined sharply during the 3rd epoch. While calculating accuracy, we considered getting either medicine name or dosage wrong to be a mistake for the entire medication.

\begin{figure}[ht]
    \centering
    \includegraphics[width=1\linewidth]{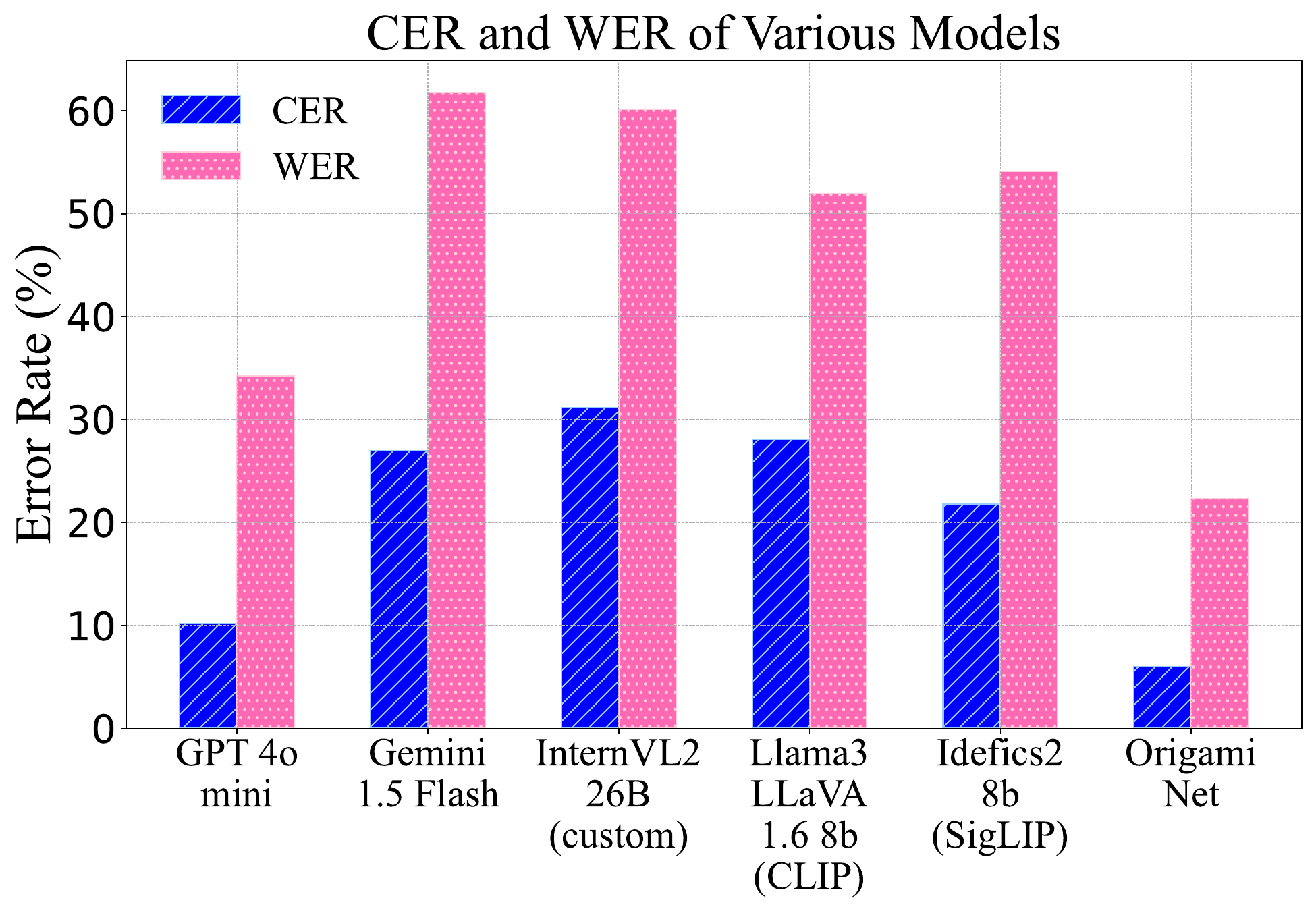}
    \caption{Error rates of various models on the IAM Line handwriting dataset (lower is better).}
    \label{fig:cer-wer}
\end{figure}

\textit{\textbf{Things to note:}} We now noticed the following:
\begin{enumerate}
    \item \textit{Importance of the Vision Encoder:} The first LLaVA paper states that the the CLIP vision encoder's output is aligned with the pre-trained LLM word embeddings via the projector \cite{liu2023llava}. Consequently, if CLIP's handwriting performance is inadequate, the LLM has limited capacity to rectify these deficiencies during fine-tuning. 
    \item \textit{Challenge with CLIP at HWR:} Although zero-shot CLIP generally outperforms fully supervised linear classifiers on ResNet-50, its performance is notably deficient on the MNIST dataset \cite{radford2021learningtransferablevisualmodels}. 
    \item \textit{The best performing model:} For our analysis, we evaluated several prominent MLLMs
    on the IAM Line handwriting dataset against the non-transformer state-of-the-art model that is OrigamiNet \cite{yousef2020origaminet}. The error rates of each model are plotted in \cref{fig:cer-wer}. While general MLLMs have demonstrated somewhat comparable performance against non-transformer state-of-the-art methods in OCR tasks, our findings reveal a far more substantial performance gap in HWR \cite{liu2023hidden, laurençon2024mattersbuildingvisionlanguagemodels}. We must highlight that HWR is generally more challenging than OCR. We proceeded to use Idefics2 due to its superior HWR performance. We must also mention that Idefics2 has been trained on PDF documents to improve its OCR performance \cite{laurençon2024mattersbuildingvisionlanguagemodels}. \\

\end{enumerate}

\textit{\textbf{Medicine names and dosages with Idefics2:}} We fine-tuned Idefics2, which employs the SigLIP vision encoder and Mistral 7B \cite{laurençon2024mattersbuildingvisionlanguagemodels, zhai2023sigmoidlosslanguageimage}. Idefics-2 supports image resolutions up to 980x980 compared to LLaVA's 672x672 and QWEN's 480x480, making it especially strong at OCR \cite{laurençon2024mattersbuildingvisionlanguagemodels}. We used QLoRA and Distributed Data Parallel (DDP) for training, and used the same rank and alpha as in LLaVA. We trained on 6 A6000s for a total of 13 days (including below experiments).

Idefics2 and LLaVA perform well on common medications but struggle with rarer ones (see \cref{fig:target-vs-pred-mix}). For Idefics2, along with no extra information for 1 epoch (\textbf{Model A}), we investigated augmenting the recognition process with: 
\begin{enumerate}
    \item Doctor's specialty (\textbf{Model B}): This helps the model narrow down likely prescriptions for each specialty. We ran a second epoch to see if scores would improve (\textbf{Model C}).
    \item Patient age, gender, and top 15 most frequently prescribed medications by the doctor for 1 epoch (\textbf{Model D}): This information helps to identify the patient type and provides a clearer understanding of the doctor’s prescribing tendencies.
\end{enumerate}

\section{Results}
\label{sec:results}
\begin{table}[ht]
    \centering
\begin{tabular}{|l|cccccc|c|} 
    \hline
    \textbf{Category} & \textbf{U} & \textbf{V} & \textbf{W} & \textbf{X} & \textbf{Y} & \textbf{Z} \\ 
        \hline
        PII                   & 25        & 32        & 39        & 46        & 44        & 44        \\ 
        \hline
        Vitals                & 2         & 17        & 26        & 33        & 34        & 43        \\ 
        \hline
        Medicine Name         & 6         & 10        & 29        & 49        & 40        & 41        \\ 
        \hline
        Medicine Properties\ \ \     & 2         & 7         & 23        & 41        & 34        & 35        \\ 
        \hline
        Diagnostic Test           & \textit{0} & 11        & 35        & 50        & 42        & 38        \\ 
        \hline
        Diagnosis             & 1         & 9         & 14        & 20        & 24        & 35        \\ 
        \hline
        Average               & 7         & 14        & 28        & 40        & 36        & 39        \\ 
        \hline
    \end{tabular}
    \caption{F1 scores (\%) for different fine-tuning strategies for LLaVA. Three models were trained for 5 epochs on 10k (U), 100k (W), and 528k (X) medical records to assess data size impact on accuracy. Improved performance with spacing from scratch on 10k samples (V) led to further fine-tuning of the normal 528k model (X) with spaced targets for two additional epochs (Y and Z).}
    \label{tab:fine_tuning_scores}
\end{table}

\begin{figure}[ht!]
    \centering
    \includegraphics[width=1\linewidth]{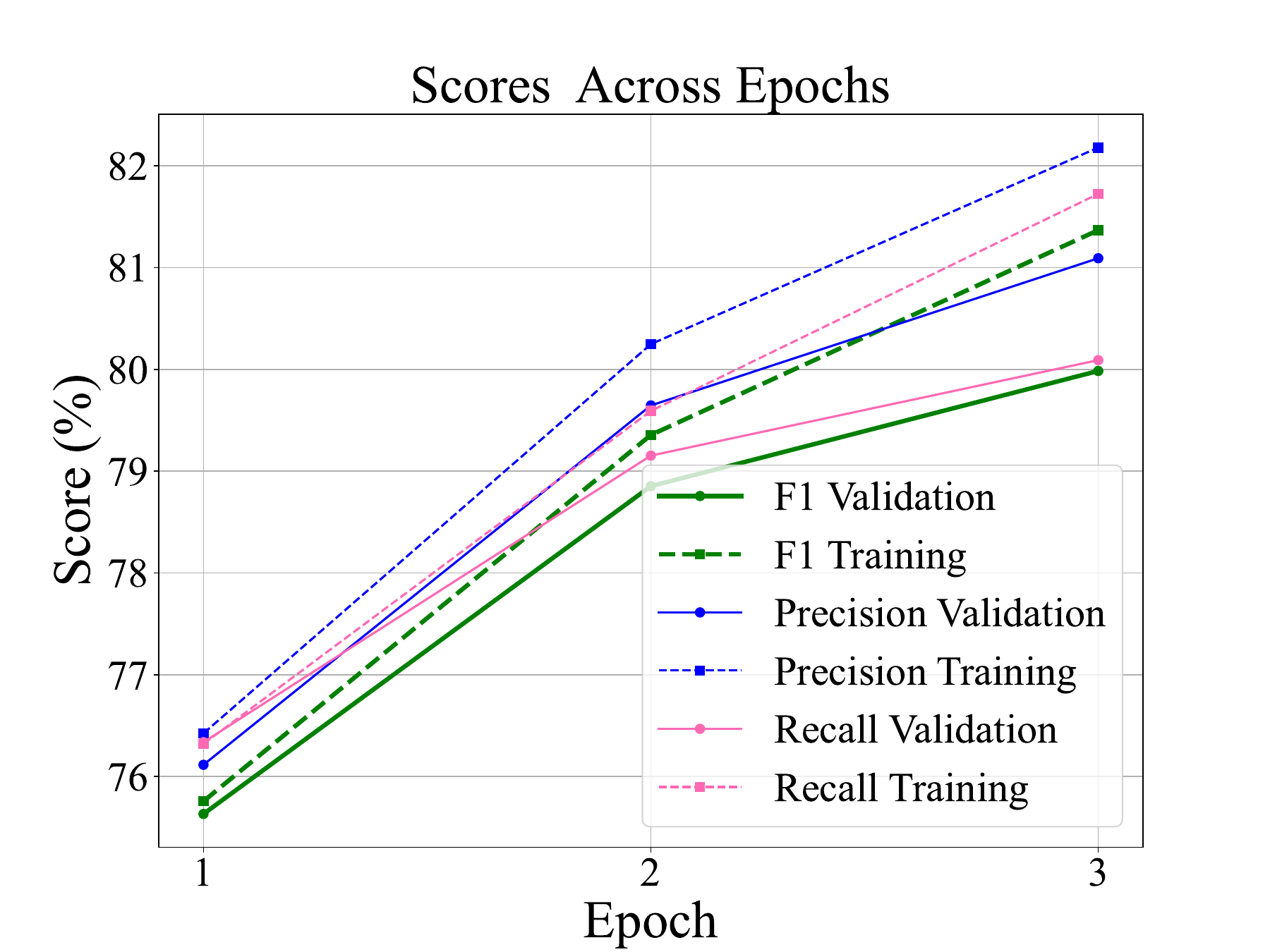}
    \caption{Training and validation scores against epochs for LLaVA. LLaVA without fine-tuning had an F1 of 2\%.}
    \label{fig:training_and_validation_accuracy_llava}
\end{figure}

\begin{figure}
    \centering
    \includegraphics[width=1\linewidth]{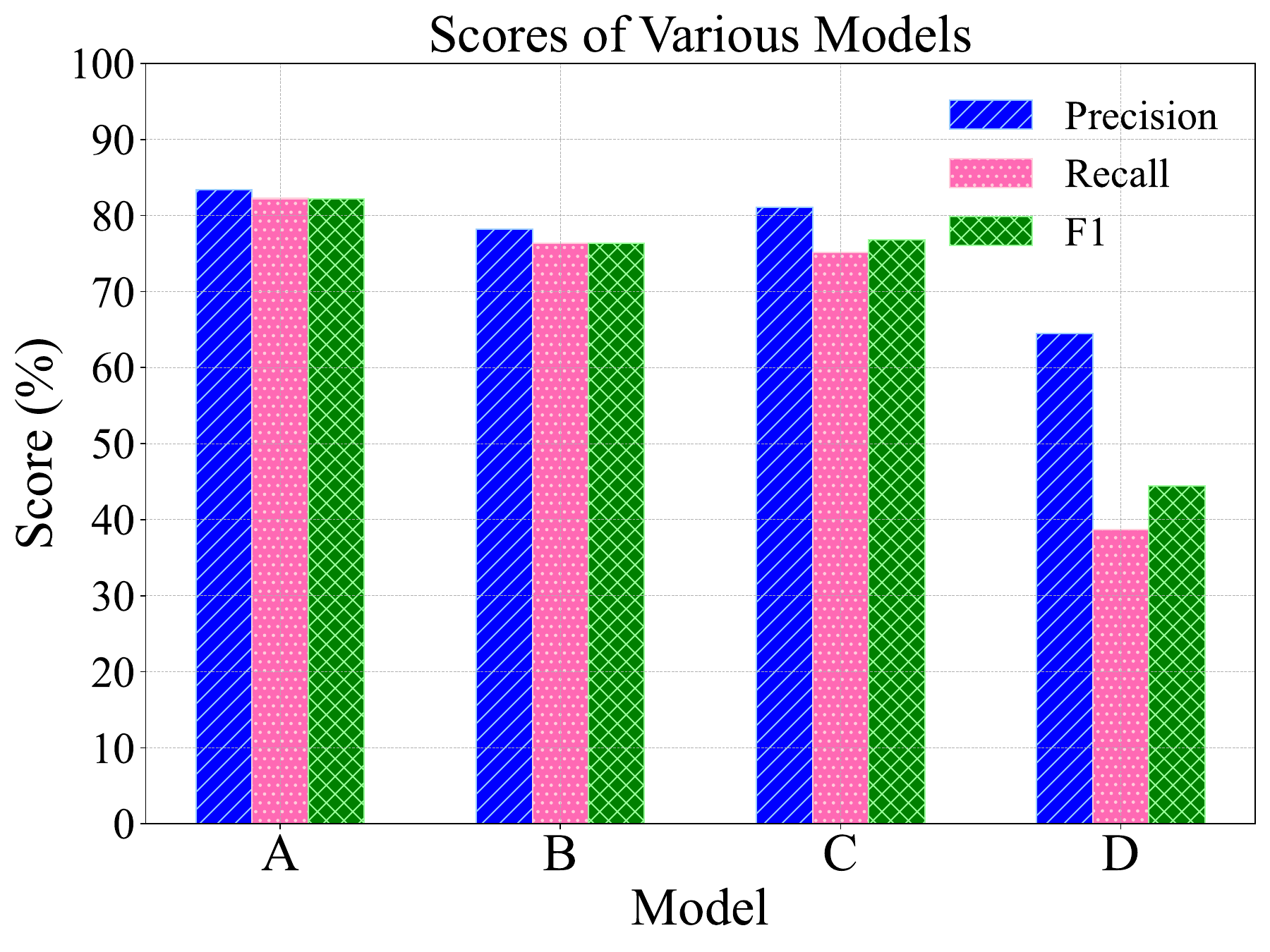}
    \caption{Scores of various models evaluated on Idefics2.}
    \label{fig:initial-ablations}
\end{figure}

\begin{figure}[ht!]
    \centering
    \includegraphics[width=1\linewidth]{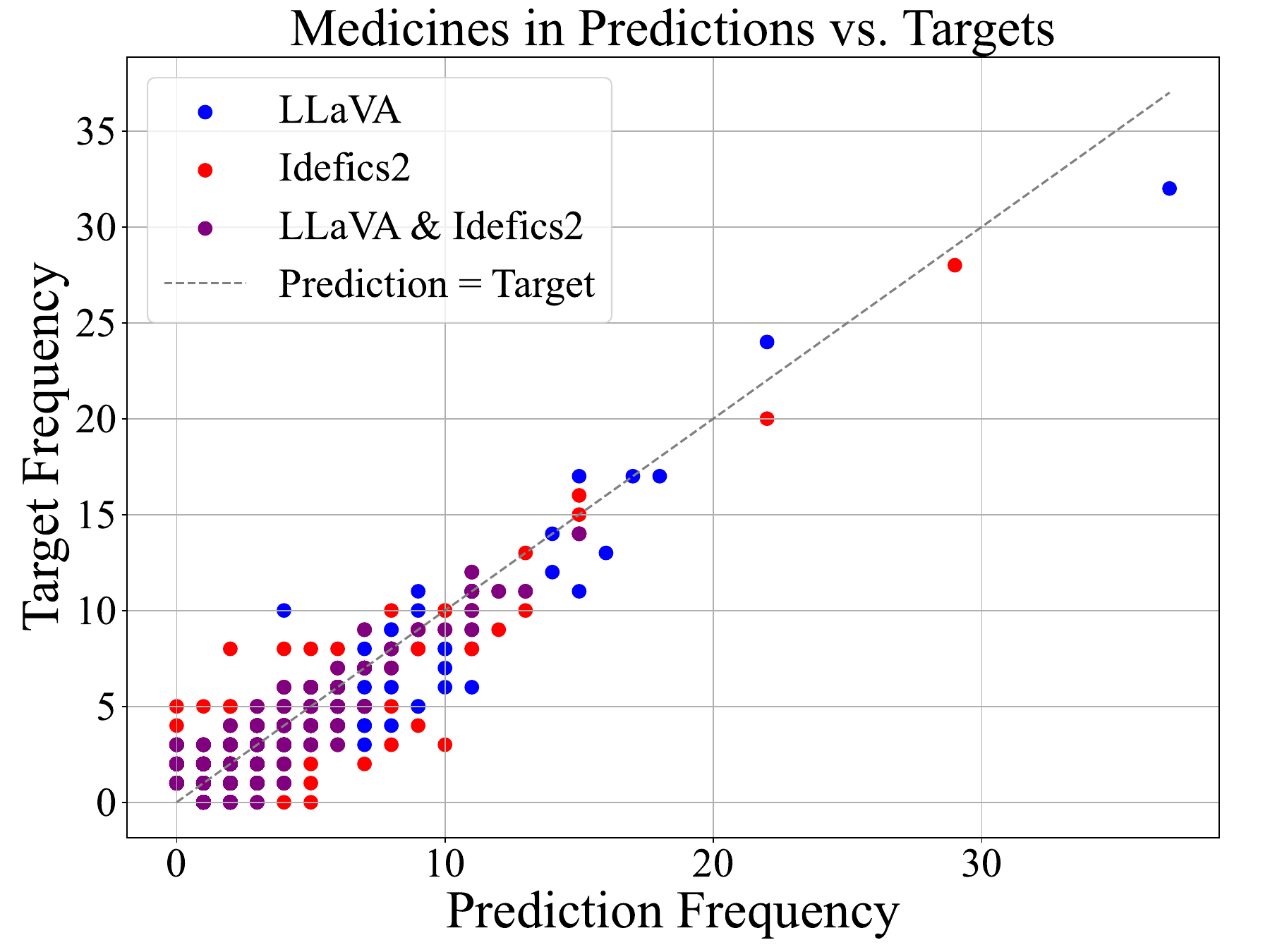}
    \caption{Comparison of predicted vs.\ target medication frequencies for 1681 medicines in 1000 records. The dotted line represents perfect accuracy. It is seen that distance from perfect accuracy increases in rarer medications. Purple points are for times when LLaVA and Idefics2 points happened to coincide.}
    \label{fig:target-vs-pred-mix}
\end{figure}

\textit{\textbf{Extracting all information:}} QWEN yielded a negligible F1 score of \textit{7\%} after 5 epochs. LLaVA achieved a maximum average F1 score of \textit{40\%}, and its accuracy on medicine names was limited to \textit{49\%}. Results are in \cref{tab:fine_tuning_scores}.

\textit{\textbf{Medicine names and dosages with LLaVA:}} We achieved a final F1 score of \textit{79.8\%}. Training and validation accuracies are illustrated in \cref{fig:training_and_validation_accuracy_llava}. Percentage of occupation of various medicines in prediction data versus target data has been plotted in \cref{fig:target-vs-pred-mix}.

\textit{\textbf{Medicine names and dosages with Idefics2:}} On our first epoch, we got an F1 score of 82\%. All our results are in \cref{fig:initial-ablations}. Percentage of occupation of various medicines in prediction data versus target data has been plotted in \cref{fig:target-vs-pred-mix}. 

\section{Discussion}
\label{sec:discussion}


Modifications to the prompts in Idefics2 did not yield performance improvements; one modification notably decreased accuracy. We attribute this outcome to the model's inherent familiarity with the prescribing patterns, given that it achieved 82\% accuracy with minimal input—solely instructions and no other details. We suggest that the model has implicitly learned information about prescribing tendencies, potentially inferring details about the doctor and patient demographics based on the medications involved as may be supported by \cref{fig:top_n}. We hypothesize that restricting the input to only the top 15 most commonly prescribed medications biased the model, causing it to overly rely on these selections. On another note, we must mention that our accuracy is in no way deployable and should not be deployed.

\section{Conclusion}
\label{sec:conclusion}

Our study demonstrates significant promise and highlights clear opportunities for improvement. With an accuracy of 82\% on a far more challenging dataset than previous works, our approach is the most practical for real-world applications yet. The ablation studies we conducted, which assess the impact of different components in the prompt, methods of representing the targets and the effect of dataset size on accuracy, provide valuable insights for future research. Furthermore, our work sheds light on the current state of Handwriting Recognition (HWR) using Large Language Models (LLMs). 



\section{Compliance with ethical standards}
\label{sec:ethics}
This study utilized a simulated dataset of medical records created by doctors. Ethical approval was not required as the research involved no interaction with human subjects.

\section{Acknowledgments}
\label{sec:acknowledgments}
This work was sponsored by Medyug Technology Pvt. Ltd., which provided the dataset and computational resources for this task. Tavish Mankash, V. S. Chaithanya Kota, and Anish De are high school students, and Tavish and Chaithanya are also interning at Medyug Technology. Praveen Prakash serves as Medyug Technology's CTO, and Kshitij Jadhav is an assistant professor at IIT Bombay.

\bibliographystyle{IEEEbib}
\bibliography{refs}

\end{document}